\documentclass[conference]{IEEEtran}
\IEEEoverridecommandlockouts
\usepackage{cite}
\usepackage{amsmath,amssymb,amsfonts}
\usepackage{algorithmic}
\usepackage{graphicx}
\usepackage{textcomp}
\usepackage{xcolor}
\usepackage{subfig}
\def\BibTex{{\rm B\kern-.05em{\sc i\kern-.025em b}\kern-.08em
    T\kern-.1667em\lower.7ex\hbox{E}\kern-.125emx}}

 \makeatletter
\newcommand{\linebreakand}{%
  \end{@IEEEauthorhalign}
  \hfill\mbox{}\par
  \mbox{}\hfill\begin{@IEEEauthorhalign}
}
\makeatother
\begin{document}

\title{Comparative Evaluation of Clustered Federated Learning Methods\\
\thanks{This work was carried out as part of the Data Mobility action of the Vilagil project, a project co-financed by "Toulouse Métropole" and by "France 2030" under "Territoires d'innovation" program operated by "La Banque des Territoires".}
}
\author{\IEEEauthorblockN{Michael Ben Ali}
\IEEEauthorblockA{\textit{IRIT, CNRS} \\ \textit{Université Toulouse, UT3}\\
Toulouse, France \\
Michael-Eddy.Ben-Ali@irit.fr}

\and
\IEEEauthorblockN{Omar El-Rifai}
\IEEEauthorblockA{\textit{IRIT, CNRS} \\ \textit{Université Toulouse, UT3} \\ 
Toulouse, France \\
Omar.El-Rifai@irit.fr}

\and
\IEEEauthorblockN{Imen Megdiche}
\IEEEauthorblockA{\textit{IRIT, CNRS} \\ \textit{INU Champollion, ISIS Castres} \\
Castres, France \\
imen.megdiche@irit.fr}
\linebreakand

\IEEEauthorblockN{André Peninou}
\IEEEauthorblockA{\textit{IRIT, CNRS} \\ \textit{Université Toulouse, UT2J} \\
Toulouse, France \\
andre.peninou@irit.fr}
\and
\IEEEauthorblockN{Olivier Teste}
\IEEEauthorblockA{\textit{IRIT, CNRS} \\ \textit{Université Toulouse, UT2J} \\
Toulouse, France \\
olivier.teste@irit.fr}
}

\maketitle

\begin{abstract}
Over recent years, Federated Learning (FL) has proven to be one of the most promising methods of distributed learning which preserves data privacy. As the method evolved and was confronted to various real-world scenarios, new challenges have emerged. One such challenge is the presence of highly heterogeneous (often referred as non-IID) data distributions among participants of the FL protocol. A popular solution to this hurdle is Clustered Federated Learning (CFL), which aims to partition clients into groups where the distribution are homogeneous.

In the literature, state-of-the-art CFL algorithms are often tested using a few cases of data heterogeneities, without systematically justifying the choices. Further, the taxonomy used for differentiating the different heterogeneity scenarios is not always straightforward.

In this paper, we explore the performance of two state-of-the-art CFL algorithms with respect to a proposed taxonomy of data heterogeneities in federated learning (FL). We work with three image classification datasets and analyze the resulting clusters against the heterogeneity classes using extrinsic clustering metrics. Our objective is to provide a clearer understanding of the relationship between CFL performances and data heterogeneity scenarios.

\end{abstract}


\section{Introduction}
Federated Learning (FL) is a distributed machine learning paradigm designed to prioritize data privacy. Originally designed \cite{b29} to train a global machine learning model across multiple clients without sharing their data, only locally trained model weights and the number of samples are shared to a server that iteratively aggregates and updates the clients' models. 

\begin{figure}[htbp]
    \centering
    \subfloat[ All clients $i$, $1\leq i \leq N$, train their local model's parameters $\omega_i$, on homogeneous data. In this scenario, the federated model parameters $\omega$ are close to the hypothetical non-federated model with parameters $\bar{\omega}$ trained on the same data but in a centralized manner]{\includegraphics[width=0.49\linewidth]{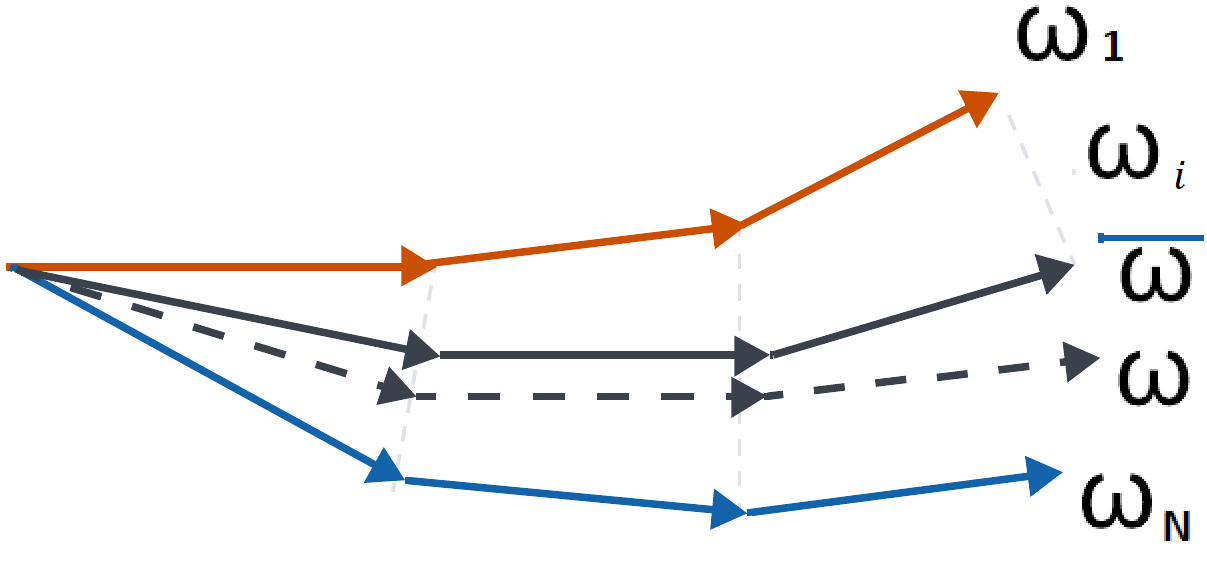}}
    \hfill
    \subfloat[ All clients $i$, $1\leq i \leq N$, train their local model's parameters $\omega_i$, on heterogeneous data. In this scenario, the federated model parameters $\omega$ drift from the non-federated model parameters $\bar{\omega}$ trained on the same data but in a centralized manner]{\includegraphics[width=0.49\linewidth]{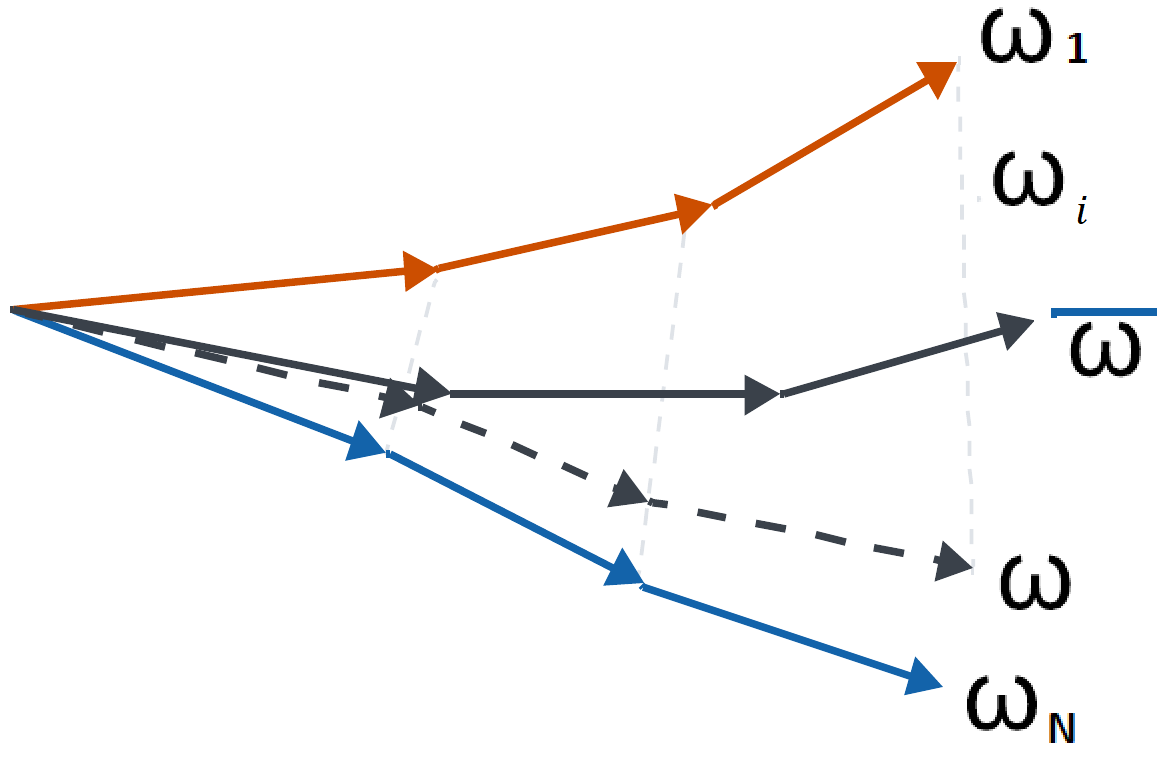}}
    \hfill
    \caption{IID scenario vs Non-IID as illustrated in \cite{b44}}
    \label{fig:drift}
\end{figure}

One of the main challenges for the convergence of FL algorithms is the presence of non-IID (non-independent and non-identically distributed) data distribution across clients. The difficulty \cite{b44} of this scenario is illustrated with Figure \ref{fig:drift}, where a comparison between the direction of parameters of a hypothetical centralized model trained on all clients' data and the parameters of the federated model is proposed in IID and non-IID scenarios. In the case of non-IID data, the gap between the global federated model's weights $\omega$ and the centralized model's weights $\bar{\omega}$ is far larger than in the case of homogeneous distributions.

The heterogeneity of client distributions has been the subject of a multitude of studies \cite{b17,b44}. In recent research, a popular solution proposed to mitigate the problem of heterogeneous client data distribution is Clustered Federated Learning (CFL) \cite{b3,b5,b9,b10,b14}. The idea behind CFL is to relax the notion of a single global model and propose multiple personalized models for particular distributions. CFL proposes a paradigm that will group clients into clusters where ideally, client data distributions are supposed to be homogeneous and where the standard FL protocol will be applied within each cluster.  It is particularly useful when we already have insight of clients' data with similar patterns.

Since a well recognized  taxonomy of data heterogeneity of client datasets for FL was proposed \cite{b17} and even if CFL has been the subject of multiple research papers, the rigorous exploration of CFL algorithms regarding each cases of heterogeneous data distribution, aiming to be comprehensive and as exhaustive as possible in covering heterogeneity cases, to our knowledge, has not been addressed yet in the literature. 

Our contribution in this paper is to rigorously study the effect of each cases of data heterogeneity on state-of-the-art CFL solutions, striving to be as exclusive as possible in each type without overlapping with other types. We propose to explore each heterogeneity types independently using three images classification datasets, MNIST, Fashion-MNIST and KMNIST.

The paper is organized as follows. Section \ref{sec:noiid} provides a description of the non-iid heterogeneous data problem in FL and introduces a data heterogeneities taxonomy.  In Section \ref{sec:CFL}, we explain in details the functionality of CFL and present two methods to classify state-of-the-art CFL solutions. In Section \ref{sec:results}, we study potential shortcomings of state-of-the-art CFL with respect to the taxonomy described in Section \ref{sec:noiid} which highlights our contribution in this paper. Finally, we conclude with some remarks and future research directions.

\section{Non-IID data distribution in Federated Learning}\label{sec:noiid}
The original state-of-the-art federated learning algorithm and aggregation method named FedAvg \cite{b29} was thought to be capable of convergence in IID and as well in non-IID settings. Unfortunately, it was proven in subsequent studies \cite{b44} that the non-IID scenario was problematic the more the distribution of clients data shifted from one another (as illustrated in Figure \ref{fig:drift}). To have a better grasp of the different cases of data heterogeneity explained in that paper, we consider the following setup.

We suppose we have $N$ clients with $i \in I = \{1,\ldots, N \}$. For each client, we consider its associate local dataset $D_i$ noted as a samples set of the form 
\begin{equation}
    D_i = \left\{(x^{(i)}_j, y^{(i)}_j) \mid 1 \leq j \leq n_i\right\}
\end{equation}

where $x=(x^{(i)}_{j})_{1 \leq j \leq n_i}$ represents the features and $y=(y^{(i)}_{j})_{1 \leq j \leq n_i}$ represents the target and with the number of local samples being $|D_i| = n_i$. 
In case of a FL supervised learning task with features $x$ and target $y$, a statistical model operates with two levels of sampling. To access a data point, we first sample a client, denoted as $i$, from the statistical distribution $P$, which represents the available clients. Then, we draw an example $(x, y)$ from the local data distribution of that client, denoted as $\mathcal{P}_i(x, y)$. When we talk about non-IID data in federated learning, we're mainly focused on the heterogeneity between the statistical distributions $\mathcal{P}_i$ and $\mathcal{P}_j$ for different clients $i$ and $j$. 

Each clients datasets $D_i$ is represented by its local joint distribution $\mathcal{P}_{i}(x, y)$ which we can factor as follows using Bayes' rule: 
\begin{equation}    
\mathcal{P}_{i}(x, y) = \mathcal{P}_{i}(y) \mathcal{P}_{i}(x | y) = \mathcal{P}_{i}(x) \mathcal{P}_{i}(y | x)
\label{eq:bayes}
\end{equation}
where $\mathcal{P}_{i}(x)$  the marginal distribution of the features (respectively $\mathcal{P}_{i}(y)$ the target) and $\mathcal{P}_{i}(y | x)$ the distribution of the target conditioned by the features (respectively $\mathcal{P}_{i}(x | y)$ the distribution of features conditioned by the target) for client $i \in I$. 
\subsection{A taxonomy of data heterogeneities in Federated Learning}\label{sec:taxonomy}

Using the notations above, we consider 5 types of data distributions heterogeneity situations \cite{b29}. We should note that those situations are in principle not mutually exclusive and two clients may encounter together any numbers of those situations.

\begin{figure}[htbp]
    \centering
    \subfloat[Concept Shift \\ on Features]{\includegraphics[width=0.09\textwidth, height=0.09\textwidth]{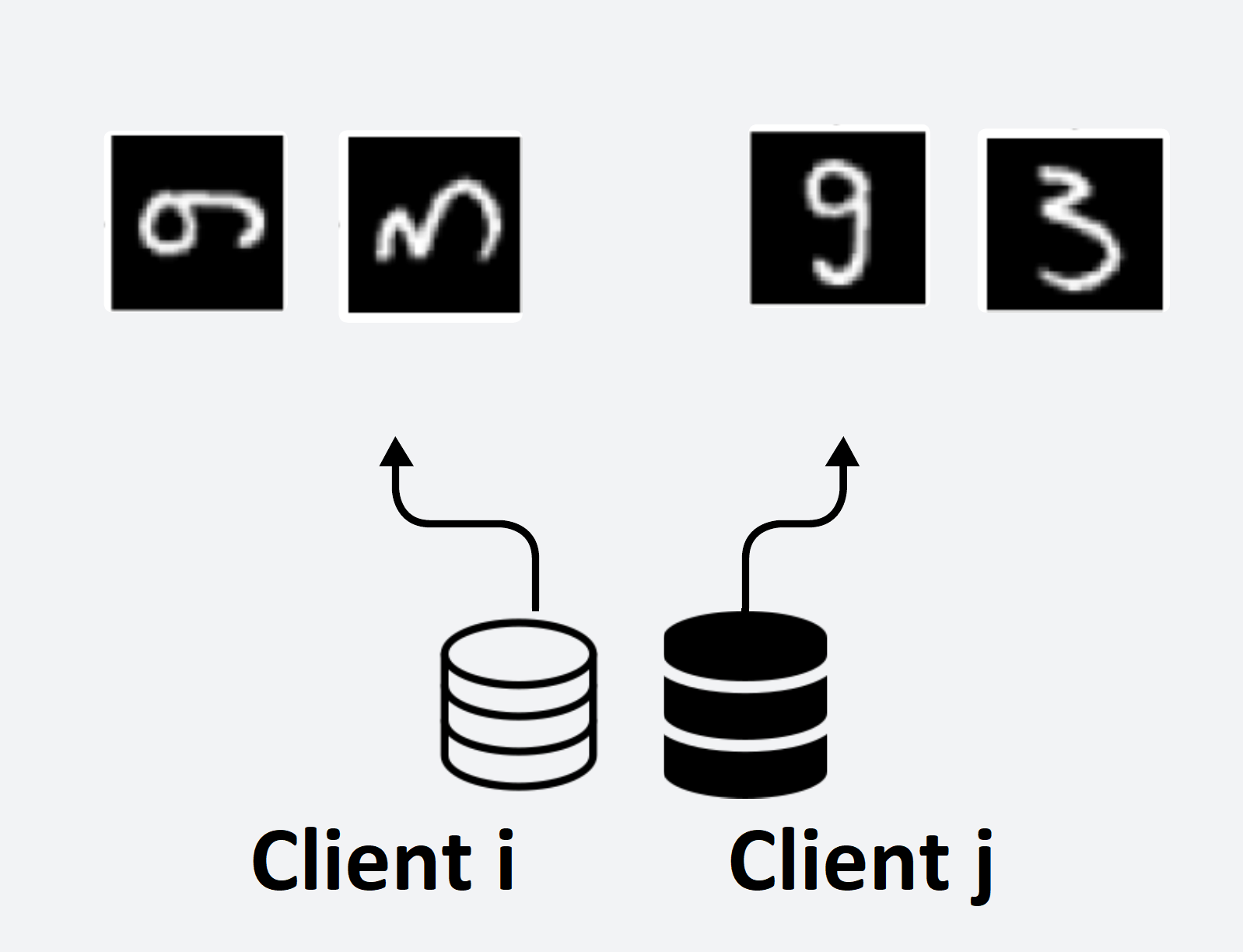}}
    \hfill
    \subfloat[Concept Shift on Labels]{\includegraphics[width=0.09\textwidth, height=0.09\textwidth]{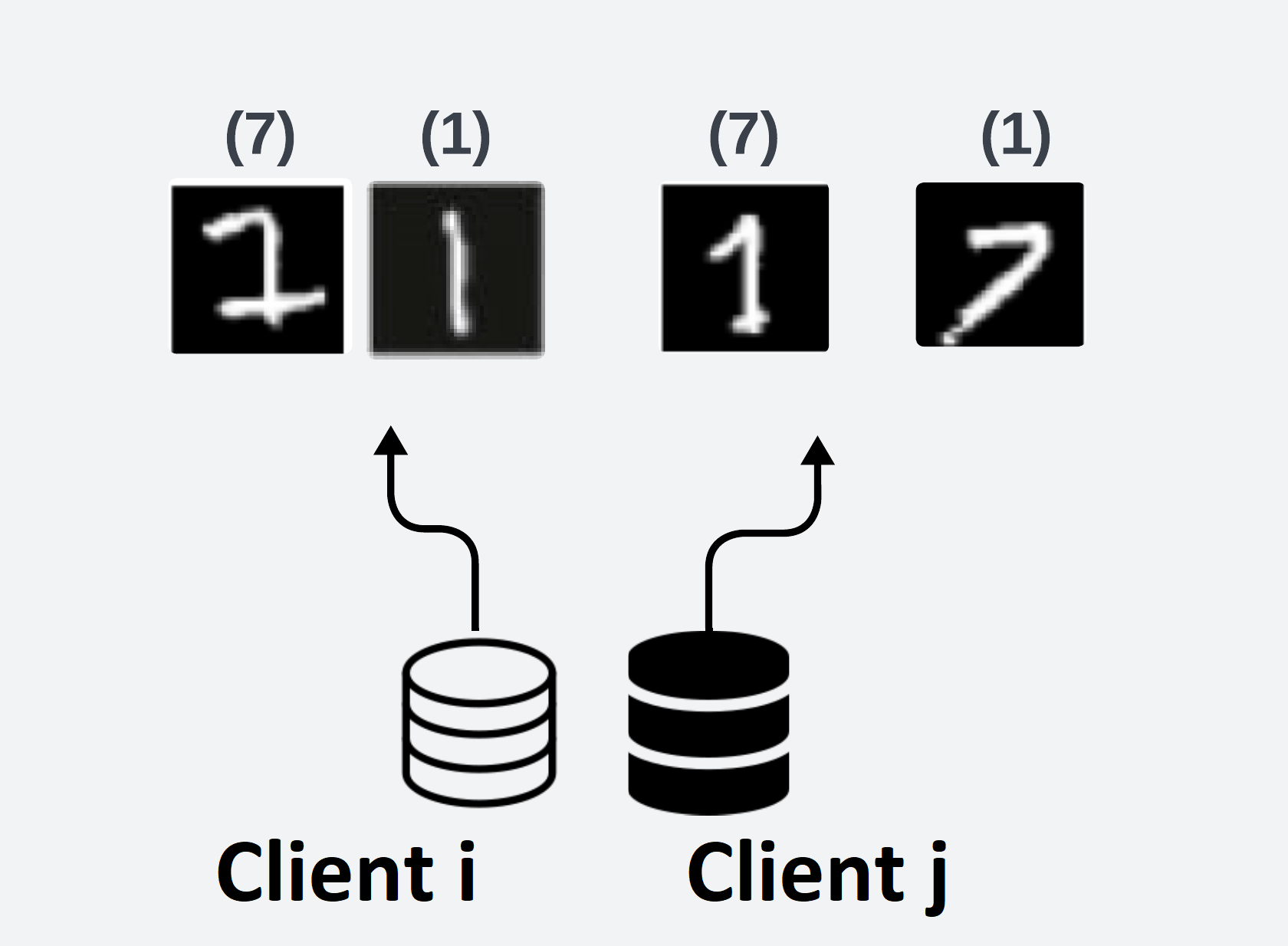}}
    \hfill
    \subfloat[Feature Distribution Skew]{\includegraphics[width=0.09\textwidth, height=0.09\textwidth]{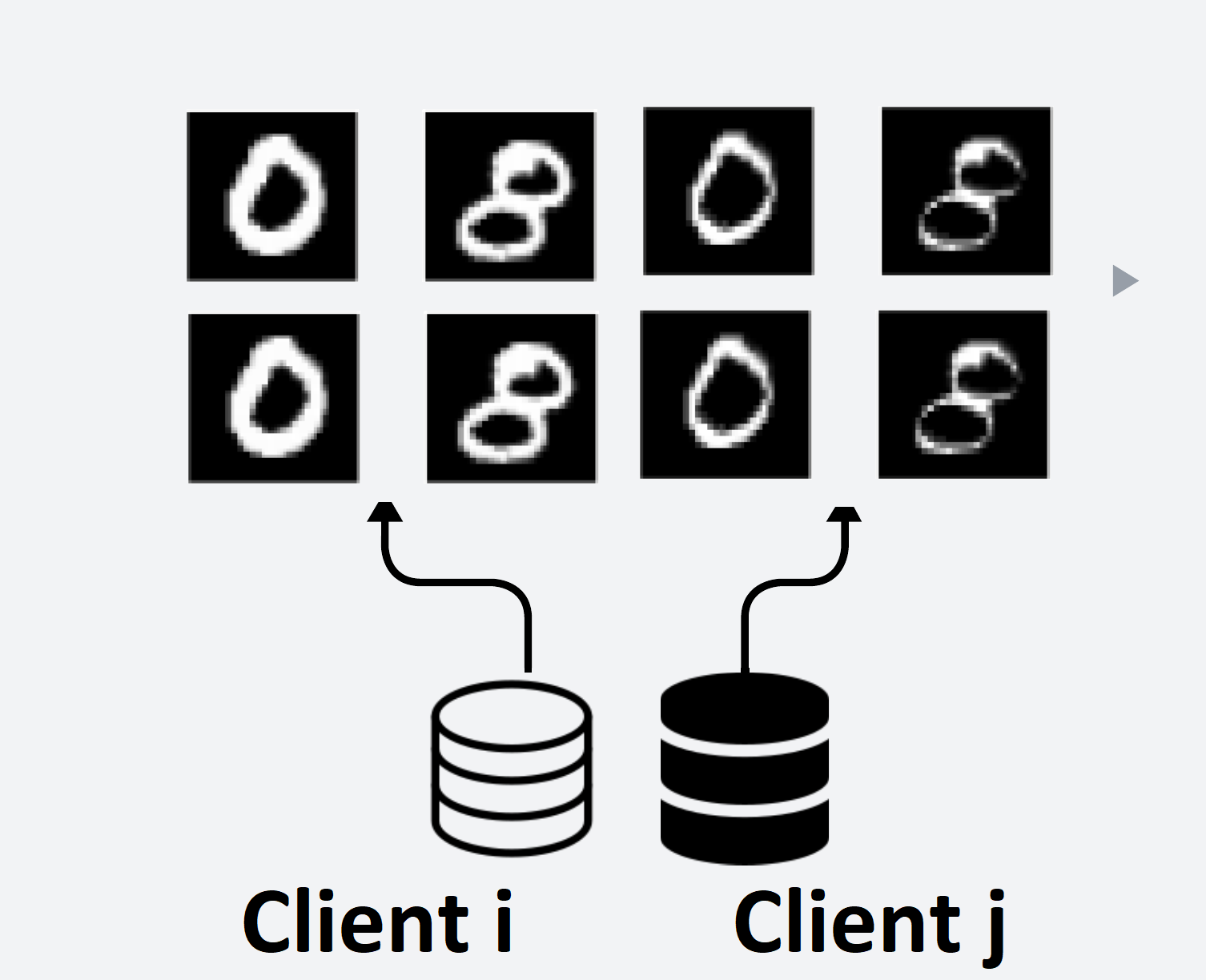}}
    \hfill
    \subfloat[Label Distribution Skew]{\includegraphics[width=0.09\textwidth, height=0.09\textwidth]{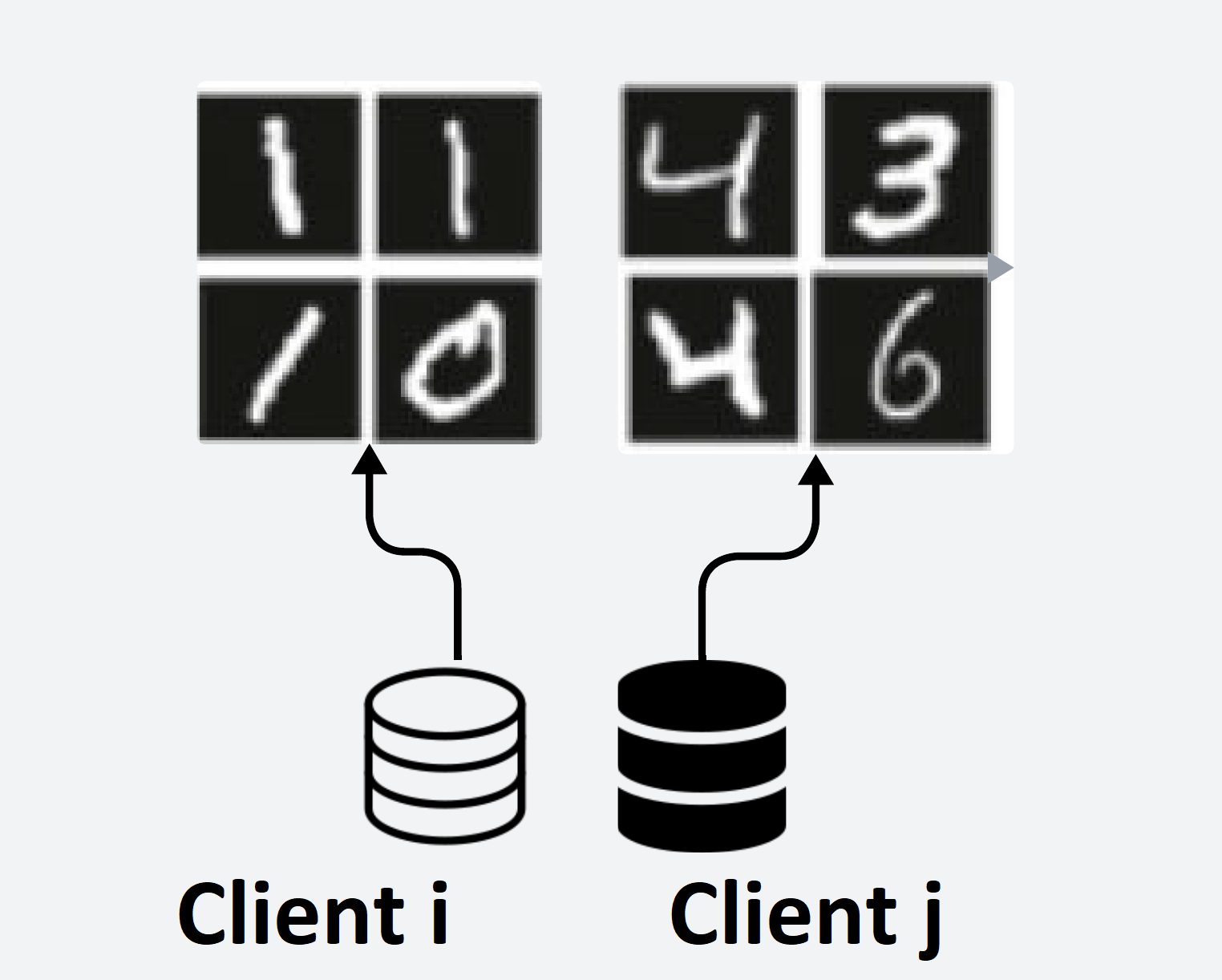}}
    \hfill
    \subfloat[Quantity Skew]{\includegraphics[width=0.09\textwidth, height=0.09\textwidth]{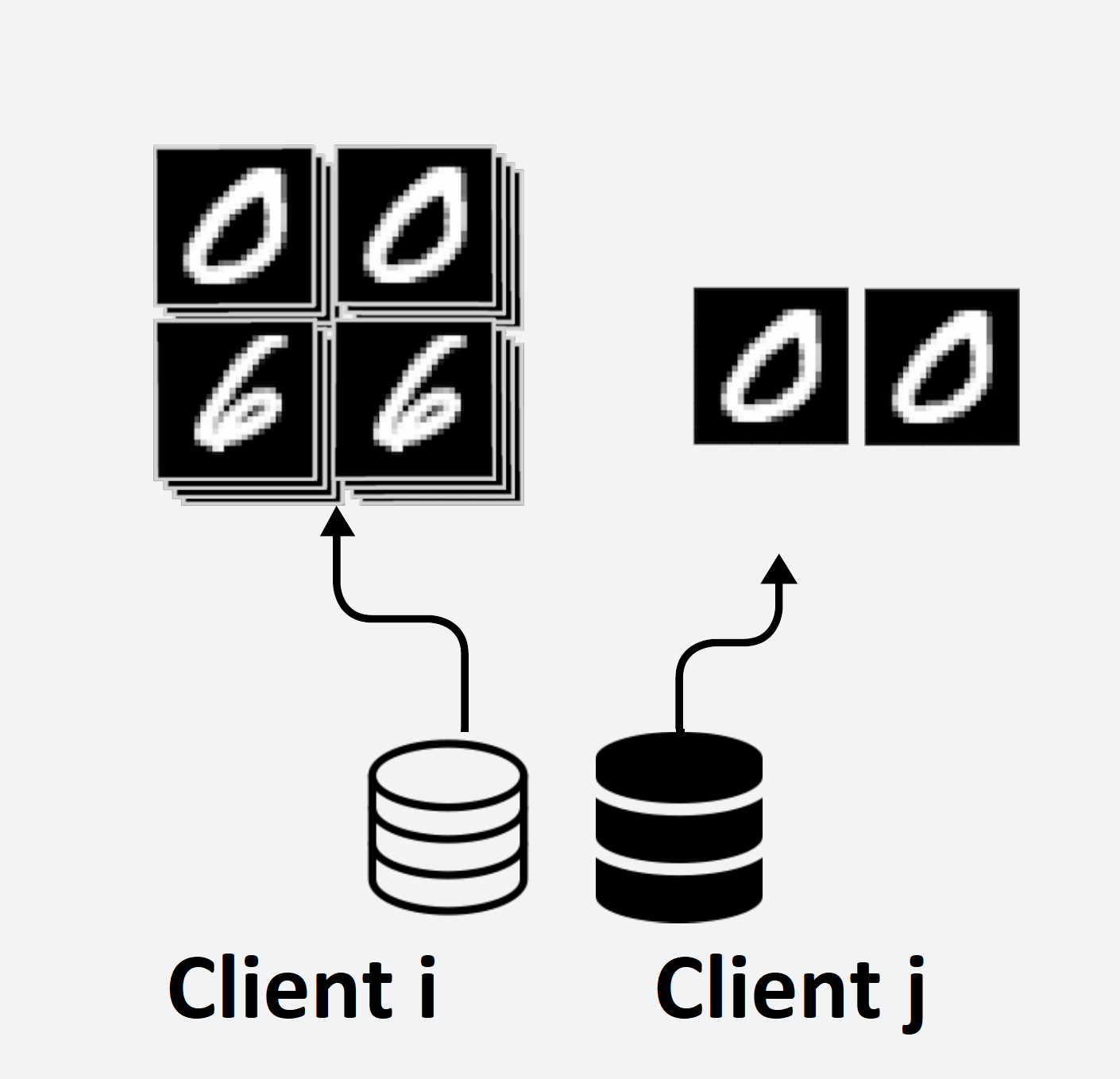}}
    \caption{Illustration of non-IID categories for two clients $i$ and $j$ with samples from the MNIST dataset.}
    \label{fig:non_iid_categories}
\end{figure}

For client $i\in I$ and client $j\in I$  with $i \neq j$, we can consider $D_i$ and $D_j$ as having heterogeneous distributions if one or multiple of the following conditions apply:
\begin{enumerate}[\IEEEsetlabelwidth{12)}]
\item \textbf{Concept shift on features}:  (same label, different features) $\mathcal{P}_i(x|y) \neq \mathcal{P}_j(x|y)$ for clients $i$ and $j$  even if $\mathcal{P}_i(y) = \mathcal{P}_j(y)$. This can be artificially attained in MNIST if we rotate the images in certain devices. (see Figure \ref{fig:non_iid_categories}.a.)

\item \textbf{Concept shift on labels}: (same features, different label) $\mathcal{P}_i(y|x) \neq \mathcal{P}_j(y|x)$ even if $\mathcal{P}_i(x) = \mathcal{P}_j(x)$. Although more likely in dataset reflecting divergence of labels concept or labels annotation error, we can illustrate such a scenario in the MNIST dataset if for instance the digits ``1'' in a user's dataset resembles  digit ``7'' in another user's or if two digits labels are swapped either by error or for malicious purposes. (see Figure \ref{fig:non_iid_categories}.b.).

\item \textbf{Feature distribution skew}: $\mathcal{P}_i(x) \neq \mathcal{P}_j(x)$ even if 
    $\mathcal{P}_i(y|x) = \mathcal{P}_j(y|x)$. In the MNIST dataset, we can artificially create this heterogeneity using different type of writing with image erosion and dilatation simulating the type of pen used. For example, we can have a dilated image simulate bold writing with a marker pen for client $i$ and eroded image with finer writing simulating a pencil writing for $j$ like in Figure \ref{fig:non_iid_categories}.c.

\item \textbf{Label distribution skew}: $\mathcal{P}_i(y) \neq \mathcal{P}_j(y)$   even if $\mathcal{P}_i(x|y) = \mathcal{P}_j(x|y)$. We can create this scenario by unevenly distributing the digits to the clients such that one client has a majority ``1''s, another a majority of ``4''s and so on as illustrated in Figure \ref{fig:non_iid_categories}.d. It corresponds to the natural cases of imbalanced datasets across clients.

\item \textbf{Quantity Skew}: the volume of data differs significantly among clients. Meaning that $|D_i|\ll|D_j|$ or $|D_j|\ll|D_i|$. (see Figure \ref{fig:non_iid_categories}.e.)        
    \label{fig:non_iid_taxonomy}

\end{enumerate}

For a clearer understanding, we illustrate in Figure \ref{fig:non_iid_categories} each type of non-iid data distribution corresponding to the taxonomy with a corresponding example from the MNIST dataset (The numbered items from the categories correspond to the letter items in the Figure in alphabetical order). 

\subsection{Related works : Datasets and heterogeneity types in CFL literature}
Even though the taxonomy defined in subsection \ref{sec:taxonomy} is well recognized across scientific papers, it is often ambiguously used. Most CFL papers do not clearly define which heterogeneity cases they are addressing. 

To rectify this, we use the taxonomy as a guideline in Table \ref{tab:datasets-noniid}, where we provide an overview of the different types of datasets, clearly identifying which types of heterogeneities, as defined by the taxonomy, are explored in CFL studies. This table reveals that none of the reviewed studies have examined the global impact of all data heterogeneity types on CFL algorithms, often focusing on seemingly random or specific cases instead. Among all the papers classified in Table \ref{tab:datasets-noniid}, only \cite{b2,b3,b5,b10,b15,b30} address two heterogeneity types out of five, while all others consider only one particular case. Moreover, it is important to note that none of these papers tested the impact of quantity skew scenarios. These observations form the foundation of our research.

We also observe that most papers focus on image classification datasets like MNIST, EMNIST, FEMNIST, Fashion-MNIST, FedCelebA and CIFAR-10, with only a few addressing other type of data like text and time series. We believe the main reason behind this is the difficulty in clearly identifying the taxonomy cases for tasks other than image classification, where the cases can be more easily defined (as illustrated by the examples derived from MNIST in subsection \ref{sec:taxonomy}).

\begin{table}[htbp]

\footnotesize
\centering
\renewcommand{\arraystretch}{1.5} 
\setlength{\tabcolsep}{4pt} 
\resizebox{0.5\textwidth}{!}{%
\begin{tabular}{|c|c|c|}
\hline
\textbf{Dataset} & \textbf{Suggested Category of non-IID situation} & \textbf{Papers} \\
\hline
MNIST & Concept shift on labels & \cite{b3,b5,b10,b36} \\
 & Concept shift on features & \cite{b5,b7,b14,b15,b30,b41} \\
 & Label distribution skew & \cite{b3,b9,b15,b22,b42} \\
\hline
EMNIST & Concept shift on features & \cite{b2,b30,b34} \\
 & Label distribution skew & \cite{b2} \\
\hline
FEMNIST & Feature distribution skew & \cite{b9, b14, b26, b30} \\
 & Concept shift on labels & \cite{b10} \\
 & Label distribution skew & \cite{b41} \\
\hline
Fashion MNIST & Concept shift on labels & \cite{b10,b27} \\
 & Label distribution skew & \cite{b15,b22} \\
 & Concept shift on features & \cite{b15} \\
\hline
FedCelebA & Feature distribution skew & \cite{b26} \\
\hline
CIFAR-10 & Concept shift on labels & \cite{b27,b30,b36} \\
 & Concept shift on features & \cite{b14,b15,b34} \\
 & Label distribution skew & \cite{b15,b30,b41,b42} \\
\hline

Shakespeare & Feature distribution skew & \cite{b7} \\
\hline
Sentiment140 & Feature distribution skew & \cite{b9} \\
\hline
ml-100K & Concept shift on labels & \cite{b25} \\
\hline
ml-1M & Concept shift on labels & \cite{b25} \\
\hline
Ag-News & Feature distribution skew & \cite{b36} \\
\hline
Time series data & Feature distribution skew & \cite{b18} \\

\hline
\end{tabular}
}
\caption{Summary of datasets and tasks tested in the CFL literature}
\label{tab:datasets-noniid}
\end{table}

\section{Clustered Federated Learning}\label{sec:CFL}
As stated in previous sections, Clustered Federated Learning is one of the most recently researched method to mitigate the negative effect of data heterogeneity across clients in FL scenario.  

In this Section, we define a generic cluster-based federated optimization formulation, and then, present the two type of solution approaches we found in the literature and classify them this way: namely, ``Server-side Clustering'' in  subsection \ref{sec:serverside} and "Client-side Clustering' in subsection \ref{sec:clientside}.

Let clients $i \in \{1,\ldots,N\}$ having datasets $D_i$ each containing $|D_i|= n_i$ data samples which are supposed to span together  $K$ different data distributions $\mathcal{P}_1,\ldots, \mathcal{P}_K$ such that $1 \leq  K \leq N$. That is, the total number of distribution is less than or equal to the number of clients.  
This implies that the clients can be partitioned into $K$ disjoint clusters $C_1,\ldots, C_K$ where the clients' datasets distributions are homogeneous intra-cluster. For the sake of simplicity in notations, each cluster will be considered as a disjoint partition of $I =\{1,\ldots,N\}$ with the notation $i\in C_k$ meaning that the client of index $i$ is inside cluster $C_k$ (i.e.: $I = \bigcup\limits_{k=1}^{K}C_k$). We would then have $K$ different clusters and for $k \in \{1,\ldots,K\}$ a corresponding objective function $F_k$ to optimize the federated model on cluster $C_k$ of the form:
\begin{equation}
\min_{\omega\in\mathbb{R}^d} F_{k}(\omega) := \sum_{i\in C_k}\frac{n_i}{\sum_{j\in C_k}{n_j}}f_{i}(\omega)
\label{eq:clustering}
\end{equation}
where $\sum_{j\in C_k}{n_j}$ correspond to the total number of samples of all clients inside cluster $C_k$ and

\begin{equation}
f_{i}(\omega)=\mathbb{E}_{(x,y) \sim D_i}[L_{i}(x,y,\omega)] \hspace{3pt} \forall \hspace{1pt} i \hspace{1pt} \in C_k
\end{equation}

that is the expected values of the expected value of the local loss function $L_i$ calculated with feature and target $(x,y)$ following the distribution of dataset $D_i$ with model of parameters $\omega\in \mathbb{R}^d$.

Based on this formulation , we describe two of the main CFL solution approaches which we found in the literature. 

\subsection{Server-Side Clustering}\label{sec:serverside}

First, and arguably the most intuitive approach as originally proposed in \cite{b13} is often referred to as gradient-based (or model-based) clustering. The studies following this approach solve a clustering problem (c.f Equation (\ref{eq:clustering})) by only using models weights. This strategy has the advantage of requiring minimum set-up over the baseline FL approach. Specifically, after a number of rounds of global FedAVG aggregations, the central server calculates clusters with k-means applied to the models weights in a "one-shot" manner and then, cluster-specific federated models are optimized separately for each cluster using the FedAVG algorithm.

Practically, in the clustering algorithm, we first define a cluster-associated representative point $\mu_k\in \mathbb{R}^d $ essential for the formation of each cluster $C_k$ for all $k \in \{1,\ldots,K\}$. The central server will determine each client $i\in \{1, \ldots,N\}$ cluster membership by minimizing the distance between its model parameters $\omega_i \in \mathbb{R}^d$ and the cluster-associated representative point. The formulation can be expressed as k-means-like optimization as follows:

\begin{equation}
\min_{\mu_k,  k \in \{1, \ldots, K\}}{\sum_{k=1}^{K}\sum_{i=1}^{N}\textbf{1}_{i \in C_k} \hspace{1pt} \text{dist}(\omega_{i},\mu_{k})} 
\label{eq:clustergrad}
\end{equation}

This approach is the baseline of many other CFL algorithm \cite{b7,b9,b15,b36} with \textbf{dist} being a generic distance function such as the originally proposed Euclidean distance between models parameters and $\textbf{1}_{i \in C_k}=1$ if client $i$ is in cluster $C_k$, 0 otherwise. Since the cluster assignment is determined by the central server, throughout the remainder of this paper, this type of clustering method will be refer to as ''Server-side'' clustering. 

\subsection{Client-Side Clustering}\label{sec:clientside}

An alternative to Server-side clustering is to assign the computational burden of determining cluster membership to client nodes. In this case, clients are presented with several possible models, initialized by the central server. The clients then evaluate the loss of their training data on these different cluster models and choose to adhere to the cluster whose model minimizes their loss. This specific approach is named IFCA (Iterative Federated Clustering Algorithm) by its author \cite{b14}. Once an initial cluster assignment has been chosen and at each communication round, each client iteratively re-evaluates and determines its optimal cluster. The clusters' models will be updated by the weighted average of corresponding members models parameters after clients' local training.

Formally, let $\mu_{k}\in \mathbb{R}^d$ be the cluster representative point of cluster $C_k$ for all $k\in\{1,\ldots,K\}$ essential for the formation of each cluster. These representative points can be initialized by the central server. Each clients $i \in \{1,\ldots,N\}$ will evaluate their training data on the $K$ different models, each initialized with parameters $\mu_k$  and update theirs models parameters $\omega_i$ with the cluster representative point which minimizes their local loss function: 
\begin{equation}
\omega_i = \underset{\mu_k,  k \in \{1, \ldots, K\}}{\arg\min}{\mathbb{E}_{(x,y) \sim D_i}[L_i(x,y,\mu_k)]}, \quad \forall i \in \{1, \dots, N\}
\label{eq:training_loss}
\end{equation}
that is the representative point $\mu_k$ that minimize the expected values of the local loss function $L_i$ calculated with feature and target $(x,y)$ following the distribution of dataset $D_i$ with model of parameters $\mu_k$.

This approach is the baseline of many other CFL algorithm \cite{b18,b26,b34} and since the cluster assignment in this case is determined by clients, throughout the remainder of this paper, this type of clustering method will be denominate as ''Client-side'' clustering. 

\section{Comparative analysis of state-of-the-art CFL solution with respect to the Non-IID Taxonomy}\label{sec:results}

In this section, we compare Client-side and Server-side CFL in non-iid scenarios. For this, we design experiments with heterogeneous distributions as defined in Section \ref{sec:taxonomy}. For simplicity, we restrict these experiments to one class of heterogeneity per client. In other words, each clients' dataset is assumed to be homogeneous in itself. Furthermore, given that the objective of CFL is to address the challenges raised by heterogeneous data distributions, we consider clusters to be well-defined if the data distributions within each cluster is homogeneous. As such, we select the initial number of clusters to match the number of data distributions across all clients. For Client-side CFL, as the final number of clusters is dictated by the algorithm, some clusters may end up with no members by the end of the process. 

We use the datasets MNIST, fashion-MNIST, and KMNIST \cite{b6} for all our experiments. All three datasets contain labeled images from 0 to 9. We assign data to clients to match the desired heterogeneity distribution as described in the following Sections. This straightforward architecture was shown to perform efficiently on the MNIST dataset using CFL \cite{b14}.

For Server-side CFL we use the algorithm in \cite{b13}, and for Client-side CFL we use the algorithm described in \cite{b14}. We choose these algorithms as they served as baselines for later improvements in the literature. We test the CFL approaches against simple federated learning on the entire dataset, and against central learning on homogeneous subsets which we call the ``oracle''. These results give us respectively a lower and upper bound on our results.

\begin{figure}[htbp]
    \centering
    \subfloat[Clients to clusters assignment which reflects the clients underlying data heterogeneity classes. In this case, all metrics equal one]{
        \includegraphics[width=0.45\linewidth]{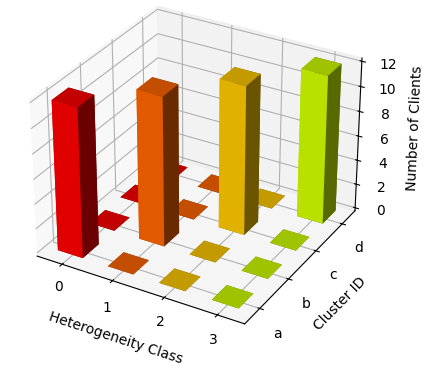}
        \label{fig:optimal_assignement}
    }
    \hfill
    \subfloat[Clients to clusters assignment which does not reflect the underlying clients' data heterogeneity classes. The resulting clustering metrics approach zero]{
        \includegraphics[width=0.45\linewidth]{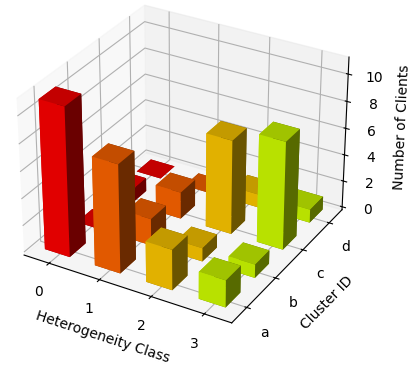}
        \label{fig:poor_assignment}
    }
    \caption{Illustration of two clients to clusters assignment edge cases}
    \label{fig:metrics_illustation}
\end{figure}

The results of our experiments are summarized in Tables \ref{tab:results_concept_shift_on_features}, \ref{tab:results_concept_shift_on_labels}, \ref{tab:results_features_distribution_skew}, \ref{tab:results_labels_distribution_skew} and \ref{tab:results_quantity_skew}. For each experiment, the tables show the dataset used, the models' accuracy, and the clustering-quality metrics obtained. The metrics reported are the Adjusted Rand Index (ARI), the Adjusted Mutual Information (AMI), as well as the homogeneity ($hom$), completeness ($cmplt$) and v-measure ($vm$) of the clusters. To illustrate what these metrics represent, we plot two edge cases in Figure \ref{fig:metrics_illustation}. In the first case (Figure \ref{fig:optimal_assignement}), the assignment follows exactly the clients' underlying data distribution with all metrics equal to one. And in the second case (Figure \ref{fig:poor_assignment}), the distribution of clients in clusters poorly reflects data heterogeneities with metrics close to zero.

All of these metrics are extrinsic measures calculated against the clustering ground truth (which is no other than the data heterogeneity classes). It is important to note that for 
CFL, the ``optimal'' clustering strategy might diverge from the strategy which yields the most efficient models. That is because Server-side CFL is agnostic to the models performances and relies on the hypotheses that data homogeneity is desirable in our scenario. This might not be true for scenarios such as ``quantiy-skew'' but analyzing these results gives us some insights on the impact of applying these algorithms without proper knowledge of our data's heterogeneity.

The following sections discuss the results for each type of heterogeneity scenario. In all our experiments, we used 48 client with 100 samples by label per client. The model used for the classification tasks is a fully connected neural network with ReLU activations \cite{b14}, with a single hidden layer of size 200  and we set the learning rate to $0.01$. For the CFL algorithms we train the models for 20 federated rounds  with each round running 10 epochs while the oracle centralized model is parameterized to 50 local epochs. The associated github used for the experiments can be found at: https://github.com/leahcimali/Comparative-Evaluation-of-Clustered-Federated-Learning-Methods

\subsection{Concept shift on features}

To introduce concept shift on features in our datasets, we create four heterogeneity classes by rotating the images 0, 90, 180, and 270 degrees. The distribution of images is such that each heterogeneity class is equally represented amongst clients. Additionally, any given client only has images of one type of rotations as explained above. As such, we set the number of clusters $k = 4$ in the CFL algorithms.

\begin{table}[ht!]
\renewcommand{\arraystretch}{1.3}
\fontsize{6.5}{8}\selectfont 
\begin{tabular}{|l|l|l|l|l|l|l|l|}
\hline
exp\_type & dataset & accuracy & ARI & AMI & hom & cmplt & vm \\ \hline
FL & fashion-mnist & 73.81 $\pm$ 0.00 & n/a & n/a & n/a & n/a & n/a \\ \hline
oracle & fashion-mnist & 83.98 $\pm$ 0.69 & n/a & n/a & n/a & n/a & n/a \\ \hline
client & fashion-mnist & 77.31 $\pm$ 2.41 & 0.48 & 0.66 & 0.5 & 1 & 0.67 \\ \hline
server & fashion-mnist & 83.66 $\pm$ 1.70 & 1 & 1 & 1 & 1 & 1 \\ \hline \hline
FL & kmnist & 72.83 $\pm$ 0.00 & n/a & n/a & n/a & n/a & n/a \\ \hline
oracle & kmnist & 87.24 $\pm$ 1.31 & n/a & n/a & n/a & n/a & n/a \\ \hline
client & kmnist & 78.44 $\pm$ 5.55 & 0.7 & 0.85 & 0.75 & 1 & 0.86 \\ \hline
server & kmnist & 86.01 $\pm$ 2.00 & 1 & 1 & 1 & 1 & 1 \\ \hline \hline
FL & mnist & 80.95 $\pm$ 0.00 & n/a & n/a & n/a & n/a & n/a \\ \hline
oracle & mnist & 92.33 $\pm$ 0.55 & n/a & n/a & n/a & n/a & n/a \\ \hline
client & mnist & 86.26 $\pm$ 4.65 & 0.7 & 0.85 & 0.75 & 1 & 0.86 \\ \hline
server & mnist & 91.78 $\pm$ 1.62 & 1 & 1 & 1 & 1 & 1 \\ \hline
\end{tabular}
\caption{Summary of results for all concept-shift-on-features experiments with average accuracy  across clusters}
\label{tab:results_concept_shift_on_features}
\end{table}

For Server-side CFL, the results show a significant improvement over standard FL, with performances comparable to the personalized oracle models across all experiments (cf Table \ref{tab:results_concept_shift_on_features}). Additionally, the clusters obtained are optimal in all Server-side experiments in the sense that they separate clients by heterogeneity perfectly. These results indicate that data heterogeneity is reflected in the models' weights which allows k-means to find the optimal clusters.   

On the other hand, with Client-side CFL, the algorithm fails to optimally cluster clients by heterogeneity, which results in performances less stellar than with server-side. Several factors can explain this. First, the initial assignment of clients to clusters is random in our experiments. This assignment can have a strong impact of the entire training process. Second, clients can be reassigned to different clusters at any point during the learning phase which leaves less room for improvement of models if the reassignment is done late in the experiment. And finally the features in this scenario very different across heterogeneity classes. This means that sub-optimal clustering will result in slightly less improvements as we see here.

From these results, It appears that personalized learning is crucial for improving performance when data features allow us to group clients with confidence such that is the case in this experiment.

\subsection{Concept shift on labels}\label{sec:concept_shift_labels}

In the case of concept shift on labels, we introduce heterogeneity by swapping some labels in the datasets. Specifically, we create 6 heterogeneity classes: swap 1 with 7, 2 with 7, 4 with 7, 3 with 8, 5 with 6, or 7 with 9. As with previous experiments, the modified datasets are distributed across clients ensuring that heterogeneity classes are equally represented.  As the resulting datasets imply diverging learning objectives, it is expected that personalized models for each heterogeneity type would also improve the models performances. This is confirmed in the results observed in Table \ref{tab:results_concept_shift_on_labels}.

\begin{table}[ht!]
\renewcommand{\arraystretch}{1.3} \fontsize{6.5}{8}\selectfont 
\begin{tabular}{|l|l|l|l|l|l|l|l|}
\hline
exp\_type & dataset & accuracy & ARI & AMI & hom & cmplt & vm \\ \hline
FL & fashion-mnist & 68.97 $\pm$ 0.00 & n/a & n/a & n/a & n/a & n/a \\ \hline
oracle & fashion-mnist & 84.48 $\pm$ 1.07 & n/a & n/a & n/a & n/a & n/a \\ \hline
client & fashion-mnist & 76.85 $\pm$ 9.33 & 0.81 & 0.92 & 0.87 & 1 & 0.93 \\ \hline
server & fashion-mnist & 84.17 $\pm$ 2.00 & 1 & 1 & 1 & 1 & 1 \\ \hline \hline
FL & kmnist & 72.21 $\pm$ 0.00 & n/a & n/a & n/a & n/a & n/a \\ \hline
oracle & kmnist & 87.98 $\pm$ 2.50 & n/a & n/a & n/a & n/a & n/a \\ \hline
client & kmnist & 76.53 $\pm$ 7.40 & 0.54 & 0.79 & 0.69 & 1 & 0.82 \\ \hline
server & kmnist & 86.44 $\pm$ 1.94 & 1 & 1 & 1 & 1 & 1 \\ \hline \hline
FL & mnist & 75.17 $\pm$ 0.00 & n/a & n/a & n/a & n/a & n/a \\ \hline
oracle & mnist & 92.17 $\pm$ 1.10 & n/a & n/a & n/a & n/a & n/a \\ \hline
client & mnist & 84.10 $\pm$ 8.89 & 0.81 & 0.92 & 0.87 & 1 & 0.93 \\ \hline
server & mnist & 91.57 $\pm$ 1.55 & 1 & 1 & 1 & 1 & 1 \\ \hline
\end{tabular}
\caption{Summary of results for all concept-shift-on-labels experiments with average accuracy  across clusters}
\label{tab:results_concept_shift_on_labels}
\end{table}

As in the previous scenario, Server-side CFL systematically reaches performances close to the oracle model's across all datasets.  Also predictably, these good performances from the Server-side CFL algorithm result from accurate clustering which is show by the perfect clustering metrics across all experiments.  

For Client-side CFL, although the clustering results are not as good, the models obtained still significantly improve on the standard FL baseline. Again This confirms that personalization can still achieve good results even if the resulting clusters are sub-optimal.

\subsection{Features distribution skew}

In the case of feature distribution skew, we introduce two types of image transformations: ``erosion'' which makes the lines thinner, and ``dilation'' which makes them bolder. We also keep unaltered images as a third  heterogeneity class. Consequently we set the  number of clusters to $k=3$. 

In this scenario, while mostly improving performance over standard FL, the CFL algorithms have a hardered time separating clients into appropriate clusters (c.f Table \ref{tab:results_features_distribution_skew}).

\begin{table}[ht!]
\renewcommand{\arraystretch}{1.3}
\fontsize{6.5}{8}\selectfont 
\begin{tabular}{|l|l|l|l|l|l|l|l|}
\hline
exp\_type & dataset & accuracy & ARI & AMI & hom & cmplt & vm \\ \hline
FL & fashion-mnist & 80.81 $\pm$ 0.00 & n/a & n/a & n/a & n/a & n/a \\ \hline
oracle & fashion-mnist & 83.83 $\pm$ 2.66 & n/a & n/a & n/a & n/a & n/a \\ \hline
client & fashion-mnist & 78.69 $\pm$ 2.45 & 0.56 & 0.73 & 0.58 & 1 & 0.73 \\ \hline
server & fashion-mnist & 82.19 $\pm$ 2.78 & 1 & 1 & 1 & 1 & 1 \\ \hline \hline
FL & kmnist & 84.49 $\pm$ 0.00 & n/a & n/a & n/a & n/a & n/a \\ \hline
oracle & kmnist & 83.65 $\pm$ 10.59 & n/a & n/a & n/a & n/a & n/a \\ \hline
client & kmnist & 75.68 $\pm$ 8.90 & 0 & 0 & 0 & 1 & 0 \\ \hline
server & kmnist & 81.61 $\pm$ 7.49 & 0.54 & 0.64 & 0.6 & 0.71 & 0.65 \\ \hline \hline
FL & mnist & 87.85 $\pm$ 0.00 & n/a & n/a & n/a & n/a & n/a \\ \hline
oracle & mnist & 88.86 $\pm$ 6.86 & n/a & n/a & n/a & n/a & n/a \\ \hline
client & mnist & 80.34 $\pm$ 9.99 & 0 & 0 & 0 & 1 & 0 \\ \hline
server & mnist & 86.40 $\pm$ 8.18 & 1 & 1 & 1 & 1 & 1 \\ \hline
\end{tabular}
\caption{Summary of results for all features-distribution-skew experiments with average accuracy  across clusters}
\label{tab:results_features_distribution_skew}
\end{table}

\begin{figure}[htp!]
\centering
\includegraphics[width=0.65\linewidth]{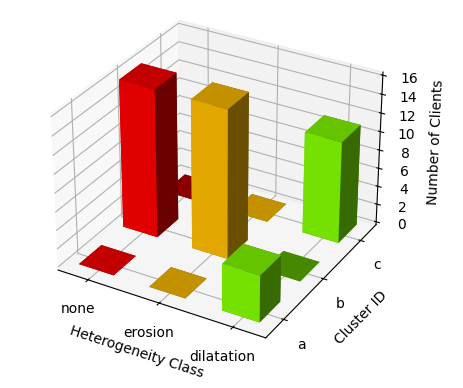}
\caption{Example clients to clusters distribution resulting from Server-side CFL using the k-mnist dataset for features distribution skew}
\label{fig:server_kmnist_fdskew}

\end{figure}

For Client-side CFL, the algorithm lumps all clients into the same cluster in 2/3 cases, resulting in poor performances. These clustering results might indicate that the loss values in the different clusters are not discriminative enough. Indeed, since the heterogeneities classes are not very different, the loss values calculated at each round may be strongly influenced by the initial random distribution of clients to clusters. Also because of the nature of the Client-side algorithm, it is possible that the cluster model with all clients have had fewer rounds to learn with the entire population of client compared to standard Fl model, which results in poorer performances.

With Server-side CFL, while we achieve model performances  consistently better than FL, the cluster structure in the `k-mnist' experiment, does not adequately separate clients by heterogeneity class. To illustrate, we see in Figure \ref{fig:server_kmnist_fdskew} that all clients with ``erosion'' or  no transformation are assigned to cluster $b$, while cluster $c$ has most clients with dilation. As opposed to the experiment with image rotations, these results suggest that the features transformations used here are not well reflected in the model weights.

\subsection{Labels distribution skew}

In the case of label distribution skew, we down-sampled some labels on each client dataset. Specifically, we create four heterogeneity classes by down-sampling 8 labels out of 10 to 0.1\% in each class. This results in selecting $k=4$ clusters for the CFL algorithms.

\begin{table}[ht!]
\renewcommand{\arraystretch}{1.3}
\fontsize{6.5}{8}\selectfont 
\begin{tabular}{|l|l|l|l|l|l|l|l|}
\hline
exp\_type & dataset & accuracy & ARI & AMI & hom & cmplt & vm \\ \hline
FL & fashion-mnist & 84.65 $\pm$ 0.00 & n/a & n/a & n/a & n/a & n/a \\ \hline
oracle & fashion-mnist & 88.04 $\pm$ 2.88 & n/a & n/a & n/a & n/a & n/a \\ \hline
client & fashion-mnist & 84.55 $\pm$ 2.88 & 1 & 1 & 1 & 1 & 1 \\ \hline
server & fashion-mnist & 87.70 $\pm$ 3.88 & 1 & 1 & 1 & 1 & 1 \\ \hline \hline
FL & kmnist & 85.64 $\pm$ 0.00 & n/a & n/a & n/a & n/a & n/a \\ \hline
oracle & kmnist & 86.66 $\pm$ 3.67 & n/a & n/a & n/a & n/a & n/a \\ \hline
client & kmnist & 82.39 $\pm$ 3.43 & 1 & 1 & 1 & 1 & 1 \\ \hline
server & kmnist & 85.89 $\pm$ 4.43 & 1 & 1 & 1 & 1 & 1 \\ \hline \hline
FL & mnist & 91.74 $\pm$ 0.00 & n/a & n/a & n/a & n/a & n/a \\ \hline
oracle & mnist & 92.61 $\pm$ 1.64 & n/a & n/a & n/a & n/a & n/a \\ \hline
client & mnist & 88.57 $\pm$ 3.30 & 1 & 1 & 1 & 1 & 1 \\ \hline
server & mnist & 92.04 $\pm$ 3.24 & 1 & 1 & 1 & 1 & 1 \\ \hline
\end{tabular}
\caption{Summary of results for all labels distribution skew experiments with average accuracy  across clusters}
\label{tab:results_labels_distribution_skew}
\end{table}

As the results in Table \ref{tab:results_labels_distribution_skew} show, both CFL algorithms consistently succeed in separating clients by heterogeneity class while the performances remain sub-optimal. The server-side algorithm reaches models' accuracies slightly better than standard FL while Client-side CFL has results similar to standard FL. The under performance of Client-side CFL algorithm can partly be explained by the low improvement margin available over the FL algorithm (as indicated by the results of the ``oracle'' model) and the iterative nature of the algorithm.


\subsection{Quantity skew}

In the case of quantity skew, we have clients with varying volumes of data, ranging from 100 percent to 20 percent of the total data distribution mentioned at the start of this section. We use this to introduce $k = 4$ heterogeneity classes, which we select as the number of clusters for CFL. Note that each client will have a balanced quantity of labels to avoid label distribution skew.

\begin{table}[ht!]
\renewcommand{\arraystretch}{1.3}
\fontsize{6.5}{8}\selectfont 
\begin{tabular}{|l|l|l|l|l|l|l|l|}
\hline
exp\_type & dataset & accuracy & ARI & AMI & hom & cmplt & vm \\ \hline
FL & fashion-mnist & 89.05 $\pm$ 0.00 & n/a & n/a & n/a & n/a & n/a \\ \hline
oracle & fashion-mnist & 85.66 $\pm$ 2.04 & n/a & n/a & n/a & n/a & n/a \\ \hline
client & fashion-mnist & 82.54 $\pm$ 4.83 & 0 & 0 & 0 & 1 & 0 \\ \hline
server & fashion-mnist & 85.21 $\pm$ 4.76 & -0.01 & -0.03 & 0.06 & 0.11 & 0.07 \\ \hline \hline
FL & kmnist & 93.44 $\pm$ 0.00 & n/a & n/a & n/a & n/a & n/a \\ \hline
oracle & kmnist & 87.35 $\pm$ 2.48 & n/a & n/a & n/a & n/a & n/a \\ \hline
client & kmnist & 86.10 $\pm$ 4.39 & 0 & 0 & 0 & 1 & 0 \\ \hline
server & kmnist & 89.71 $\pm$ 4.17 & 0 & -0.01 & 0.07 & 0.21 & 0.1 \\ \hline \hline
FL & mnist & 95.72 $\pm$ 0.00 & n/a & n/a & n/a & n/a & n/a \\ \hline
oracle & mnist & 92.07 $\pm$ 0.56 & n/a & n/a & n/a & n/a & n/a \\ \hline
client & mnist & 90.33 $\pm$ 3.76 & 0 & 0 & 0 & 1 & 0 \\ \hline
server & mnist & 92.13 $\pm$ 3.50 & 0 & -0.01 & 0.07 & 0.21 & 0.1 \\ \hline
\end{tabular}
\caption{Summary of results for all quantity skew experiments with average accuracy  across clusters}
\label{tab:results_quantity_skew}
\end{table}

It is expected that clustered federated learning with quantity skew  will fail at improving performances over standard FL. Understandably, if the objective of clustering by heterogeneity class is respected, some clusters with very low volumes of data will be constituted. These low data volume clusters are expected to perform poorly. And indeed, as we see in Table \ref{tab:results_quantity_skew}, the results validate this intuition in that CFL algorithms do not improve over a simple FL model (We note that this is also the case for the oracle algorithm given that we separate clients by heterogeneity class as well).

 It is important to note that in the standard FL model, FedAVG aggregation takes into account the number of local samples of each client. As such, clients with larger data will have a greater impact on the overall global model and potentially mitigate the discrepancy of dataset sizes among clients.

In our Server-side experiments, the clusters have no discernible logic and create models which slightly underperform. Alternatively, with Client-side CFL, the algorithm clusters all clients together which results in a set-up and performances similar to that of the baseline FL method but with less training rounds. 

\subsection{Impact of the number of clusters}

All the experiments presented assume that the number of clusters is well defined. That is, the number of clusters is always set to the number of data heterogeneity classes in the dataset. In real-life, knowing the number of heterogeneity classes in advance might not always be possible. As such, even an educated guess of an expert might miss the mark on the exact number of data heterogeneities present. To evaluate the impact of this variable, we tested the impact of the number of clusters on the accuracy for the Server-side ``concept-shift-on-labels'' problem with the mnist dataset.     

\begin{figure}[htp!]
\centering
\includegraphics[width=0.65\linewidth]{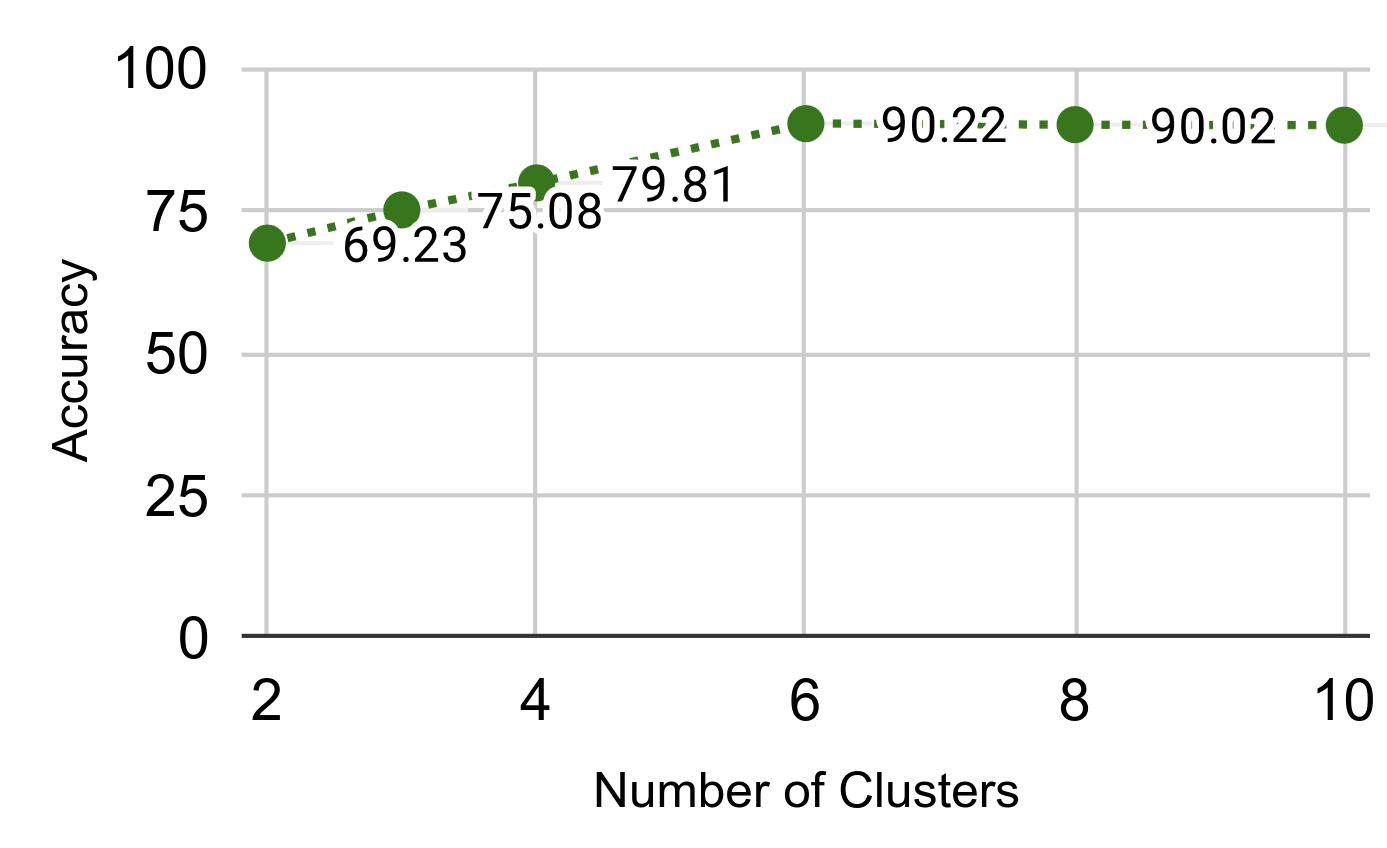}
\caption{Impact of the number of clusters on the CFL results }
\label{fig:impact_num_clusters}

\end{figure}

As we see in Figure \ref{fig:impact_num_clusters}, when the number of clusters is very low compared to the number of heterogeneity classes (here being six like in Section \ref{sec:concept_shift_labels}), the results approach those observed with the baseline FL algorithm. As we increase the number of clusters, the accuracy increases as well. We note that the accuracy seems to stabilize if we increase the number of clusters beyond the nominal value. This means that in some circumstances (and assuming sufficient data volumes per clusters), models personalization can go beyond the number of heterogeneity classes without penalizing model performances.

\section{Conclusion}\label{sec:conclusion}
In this study, we explored the performance of Server-side and Client-side Clustered Federated Learning (CFL) algorithms under systematized data heterogeneity conditions. Our findings show improved performances for several use-cases with more consistent results from server-side CFL. 

In most cases, the accuracy improvements follow a clustering stage which successfully separates clients by heterogeneity class.  This is the case in scenarios where the features or the labels are significantly different across clients such as in concept-shift-on-features and concept-shift-on-labels scenarios. On the other hand, when the differences across the heterogeneity classes are more subtle (e.g features distribution skew), even server-side CFL has a harder time separating by heterogeneity class but personalization still improves results. In the case of labels distribution skew, it seems that the heterogeneity does not have a strong impact on the model performances with standard FL achieving quite high accuracies. As such CFL method only slightly improve these results. Finally, when the data heterogeneity leads to clusters with incomplete data for learning (e.g quantity skew) it seems undesirable to use CFL at all. 

This might suggest that an informed understanding of the learning environment might be key to making adequate decisions on when and how to use CFL. Fortunately, even in the quantity-skew scenario, the performances of the models were not strongly penalized by clustering. Further, in Client-side clustering, as the initial assignment of clients to clusters has a strong impact on the entire training process, it is important to make an informed decision at this initial stage. We confirmed this intuition by starting the CFL experiments with a semi-random assignment (essentially simulating an expert opinion on the heterogeneity class of the client) and obtained improved results which approached those of server-side CFL. Further investigation in this direction is needed.


We also investigated the impact of the number of clusters on the performance of our models. We observed that, as expected, the best results are obtained when the number of clusters is close or equal to the number of data heterogeneity classes. It also appears safer to overestimate the number of heterogeneity classes (up to a certain limit) which might be explained by unaccounted heterogeneities in the data. 

Further research is needed to see if these results scale beyond simple datasets, and with the presence of a combination of data heterogeneities by client.


\vspace{12pt}


\begin{thebibliography}{00}

\bibitem{b2} A. Augello, G. Falzone, and G. Lo Re, ``DCFL: Dynamic clustered federated learning under differential privacy settings,'' in \textit{2023 IEEE International Conference on Pervasive Computing and Communications Workshops and other Affiliated Events (PerCom Workshops)}, 2023

\bibitem{b3} C. Briggs, Z. Fan, and P. Andras, ``Federated learning with hierarchical clustering of local updates to improve training on non-iid data,'' in \textit{2020 International Joint Conference on Neural Networks (IJCNN)}, IEEE, 2020.

\bibitem{b5} F. E. Castellon, A. Mayoue, J.-H. Sublemontier, and C. Gouy-Pailler, ``Federated learning with incremental clustering for heterogeneous data,'' in \textit{2022 International Joint Conference on Neural Networks (IJCNN)}, IEEE, 2022.

\bibitem{b6} T. Clanuwat, M. Bober-Irizar, A. Kitamoto, A. Lamb, K. Yamamoto and D. Ha
``Deep Learning for Classical Japanese Literature'', \textit{arxiv preprint arxiv:1812.01718}, 2018

\bibitem{b7} D. K. Dennis, T. Li, and V. Smith, ``Heterogeneity for the win: One-shot federated clustering'', \textit{International Conference on Machine Learning}, PMLR, 2021.

\bibitem{b9} M. Duan, et al., ``Fedgroup: Efficient federated learning via decomposed similarity-based clustering,'' in \textit{2021 IEEE Intl Conf on Parallel \& Distributed Processing with Applications, Big Data \& Cloud Computing, Sustainable Computing \& Communications, Social Computing \& Networking (ISPA/BDCloud/SocialCom/SustainCom)}, IEEE, 2021.

\bibitem{b10} M. Duan, et al., ``Flexible clustered federated learning for client-level data distribution shift,'' \textit{IEEE Transactions on Parallel and Distributed Systems}, 2021.

\bibitem{b13} A. Ghosh, J. Hong, D. Yin, and K. Ramchandran, ``Robust Federated Learning in heterogeneous environments,'' \textit{arXiv preprint arXiv:1906.06629}, 2019.

\bibitem{b14} A. Ghosh, J. Chung, D. Yin, K. Ramchandran, and A. Chandra, ``An efficient framework for clustered federated learning,'' \textit{Advances in Neural Information Processing Systems} 2020.

\bibitem{b15} B. Gong, T. xing, Z. Liu, W. xi, and X. Chen, ``Adaptive client clustering for efficient federated learning over non-iid and imbalanced data,'' \textit{IEEE Transactions on Big Data}, 2022.

\bibitem{b17} P. Kairouz, et al., ``Advances and open problems in federated learning,'' \textit{Foundations and Trends in Machine Learning}, 2021

\bibitem{b18} Y. Kim, et al., ``Dynamic clustering in federated learning,'' in \textit{ICC 2021-IEEE International Conference on Communications}, IEEE, 2021.

\bibitem{b22} C. Li, G. Li, and P. K. Varshney, ``Federated learning with soft clustering,'' \textit{IEEE Internet of Things Journal}, 2021.

\bibitem{b25} T. Liang et al., ``Efficient one-off clustering for personalized federated learning,'' \textit{Knowledge-Based Systems}, 2023.

\bibitem{b26} G. Long, et al., ``Multi-center federated learning: clients clustering for better personalization,'' \textit{World Wide Web}, 2023.

\bibitem{b27} Y. Luo, x. Liu, and J. xiu, ``Energy-efficient clustering to address data heterogeneity in federated learning,'' in \textit{ICC 2021-IEEE International Conference on Communications}, IEEE, 2021.

\bibitem{b29} B. McMahan, E. Moore, D. Ramage, S. Hampson, and B. A. Arcas, ``Communication-efficient learning of deep networks from decentralized data,'' in \textit{Artificial Intelligence and Statistics}, PMLR, 2017.

\bibitem{b30} M. Mehta, and C. Shao, ``A greedy agglomerative framework for clustered federated learning,'' \textit{IEEE Transactions on Industrial Informatics}, 2023.

\bibitem{b34} Y. Ruan, and C. Joe-Wong, ``Fedsoft: Soft clustered federated learning with proximal local updating,'' in \textit{Proceedings of the AAAI Conference on Artificial Intelligence}, 2022.

\bibitem{b36} F. Sattler, K.-R. Müller, and W. Samek, ``Clustered federated learning: Model-agnostic distributed multitask optimization under privacy constraints,'' \textit{IEEE Transactions on Neural Networks and Learning Systems}, 2019.

\bibitem{b41} J. Wolfrath, N. Sreekumar, D. Kumar, Y. Wang, and A. Chandra, ``Haccs: heterogeneity-aware clustered client selection for accelerated federated learning,'' in \textit{2022 IEEE International Parallel and Distributed Processing Symposium (IPDPS)}, IEEE, 2022.

\bibitem{b42} Y. Xiao, J. Shu, x. Jia, and H. Huang, ``Clustered federated multi-task learning with non-iid data,'' in \textit{2021 IEEE 27th International Conference on Parallel and Distributed Systems (ICPADS)}, IEEE, 2021.

\bibitem{b43} M. Ye, X. Fang, B. Du, P. C. Yuen, and D. Tao, ``Heterogeneous federated learning: State-of-the-art and research challenges'', \textit{ACM Computing Surveys}, 2023.

\bibitem{b44} Y. Zhao, et al., ``Federated learning with non-iid data,'' \textit{arxiv preprint arxiv:1806.00582}, 2018.


\end{thebibliography}
\end{document}